\documentclass[sigconf]{acmart}

\DeclareMathDelimiter{(}{\mathopen} {operators}{"28}{largesymbols}{"00}
\DeclareMathDelimiter{)}{\mathclose}{operators}{"29}{largesymbols}{"01}

\usepackage[hyperref]{}
\usepackage{latexsym}
\usepackage{times}
\usepackage{algorithm}
\usepackage{algpseudocode}
\usepackage{graphicx}
\usepackage{hhline}
\usepackage{multirow}
\usepackage{booktabs}
\usepackage[normalem]{ulem}
\usepackage{amsmath}
\usepackage{float}
\usepackage{caption}
\algrenewcommand{\Return}{\State\algorithmicreturn~}

\usepackage{microtype}
\AtBeginDocument{%
  \providecommand\BibTeX{{%
    \normalfont B\kern-0.5em{\scshape i\kern-0.25em b}\kern-0.8em\TeX}}}

\copyrightyear{2022}
\acmYear{2022}
\setcopyright{acmlicensed}
\acmConference[CODS-COMAD 2022]{5th Joint International Conference on Data Science \& Management of Data (9th ACM IKDD CODS and 27th COMAD)}{January 8--10, 2022}{Bangalore, India}
\acmBooktitle{5th Joint International Conference on Data Science \& Management of Data (9th ACM IKDD CODS and 27th COMAD) (CODS-COMAD 2022), January 8--10, 2022, Bangalore, India}
\acmPrice{15.00}
\acmDOI{10.1145/3493700.3493705}
\acmISBN{978-1-4503-8582-4/22/01}

\begin{document}
\sloppy
\title{AdvCodeMix: Adversarial Attack on Code-Mixed Data}

\author{Sourya Dipta Das}
\email{dipta.juetce@gmail.com}
\affiliation{%
  \institution{Jadavpur University}
  \city{Kolkata}
  \state{West Bengal}
  \country{India}
}

\author{Ayan Basak}
\email{ayanbasak13@gmail.com}
\affiliation{%
  \institution{Jadavpur University}
  \city{Kolkata}
  \state{West Bengal}
  \country{India}
}

\author{Soumil Mandal}
\email{soumil.mandal@utdallas.edu}
\affiliation{%
  \institution{University of Texas at Dallas}
  \city{Texas}
  \country{USA}}

\author{Dipankar Das}
\email{dipankar.das@jadavpuruniversity.in}
\affiliation{%
  \institution{Jadavpur University}
  \city{Kolkata}
  \state{West Bengal}
  \country{India}
}







\begin{abstract}
Research on adversarial attacks are becoming widely popular in the recent years. One of the unexplored areas where prior research is lacking is the effect of adversarial attacks on code-mixed data. Therefore, in the present work, we have explained the first generalized framework on text perturbation to attack code-mixed classification models in a black-box setting. We rely on various perturbation techniques that preserve the semantic structures of the sentences and also obscure the attacks from the perception of a human user. The present methodology leverages the importance of a token to decide where to attack by employing various perturbation strategies. We test our strategies on various sentiment classification models trained on Bengali-English and Hindi-English code-mixed datasets, and reduce their F1-scores by nearly 51\% and 53\% respectively, which can be further reduced if a larger number of tokens are perturbed in a given sentence. 
\end{abstract}




\maketitle

\section{Introduction}

In the age of globalization, code-mixed text inputs are a common phenomenon since it is quite natural for bilinguals to switch back and forth between two languages while communicating, both in verbal and textual forms~\cite{sridhar1980syntax}. Such textual instances are challenging to process as they often combine grammatical bases of both the languages in a single sentence, make use of colloquial terms and short-forms extensively, and standardized rules aren't followed while transliterating when the pair of languages have different scripts. We introduce a three-step attack strategy that can be used for generating adversarial examples using minimal resources for any type of code-mixed data (with and without transliteration). We have used our framework to evaluate the success of adversarial attacks on a few sentiment classification models~\cite{patwa2020semeval,zhang2015character,pires2019multilingual} that have been diagnosed effective on code-mixed data. Research on adversarial techniques has become an important aspect, especially for security-critical applications, as it helps us in both analyzing the fallacies of the models, and make them more robust. Some of the popular methods towards building robust pipelines include adversarial training~\cite{madry2017towards} of the model and rejection of adversarial inputs~\cite{meng2017magnet}.  
\newline
In our approach, we do not make any replacements in the original sentence based on word synonyms as observed in previous adversarial attack approaches (\citet{jin2020bert}, \citet{li2020bert}), or any such triggers at the word or sentence level~\cite{sun2020natural}, which ensures that the semantic similarity is automatically preserved. We only make phonetic perturbations at the sub-word and word-level, and replace words with their corresponding transliterated counterpart in the code-mixed sentence, as it can deceive the model in several cases despite keeping the overall structure intact. The main contributions of our paper are (a) preparation of a generalized model agnostic framework for generating adversarial examples of code-mixed data, (b) proposed novel language specific perturbation techniques that preserve the semantic structure of a sentence, (c) achieve a successful attack in a short time span per sentence. 

\section{Related Work}
Sun et al.~\cite{sun2020natural} had systematically studied backdoor attacks on text data. They have shown the impact that different trigger mechanisms like sentence, character and word level triggers, non-natural triggers and special triggers can have on the attack framework. Jin et al.~\cite{jin2020bert} had proposed the TEXTFOOLER framework that can generate adversarial text for binary text classification and text entailment tasks and have achieved state-of-the-art results on powerful models like pre-trained BERT, convolutional and recurrent neural networks. 
A novel and interesting approach, BERT-Attack, was also proposed~\cite{li2020bert} where the pre-trained BERT model was used to effectively attack fine-tuned BERT models, as well as traditional LSTM based deep learning models. Liu et al.~\cite{liu2020kk2018} have applied the concept of transfer learning using the ERNIE framework, along with adversarial training using a multilingual model, to their work. Tan et al.~\cite{tan2021code} have proposed two strong black-box adversarial attack frameworks-one word-level and another phrase-level, the latter one being being particularly effective on XNLI.
Ren et al.~\cite{ren2019generating} had proposed a novel greedy algorithm, the probability weighted word saliency (PWWS), that is based on a synonym substitution strategy. 
Li et al.~\cite{li2018textbugger} had proposed the TEXTBUGGER framework that outperforms current state-of-the-art adversarial attack frameworks in terms of attack success rate. 
Gao et al.~\cite{gao2018black} had proposed a framework to generate adversarial text in a black-box setting - DeepWordBug which can effectively generate small perturbations in the most critical tokens based on novel scoring strategies for adversarial attack.
\section{Problem and Challenges}
In this problem, we have a set of $N$ sentences $X = \{x_{1}$, $x_{2}$, ..., $x_{N}$\} with an associated set of N labels  $Y = \{y_{1}$, $y_{2}$, ..., $y_{N}$\} where total number of classes is $M$. For a given pre-trained model, $F$ needs to do a mapping $F: X \mapsto Y$ from an input sample, $x \in X$ to a ground truth label, $y_{true} \in Y$. Now, for an input sentence, $x \in X$, a valid adversarial example, $x_{adv}$ must meet the following criteria.
\setlength{\belowdisplayskip}{4pt} \setlength{\belowdisplayshortskip}{4pt}
\setlength{\abovedisplayskip}{4pt} \setlength{\abovedisplayshortskip}{4pt}
\begin{align}
\small
    F(x) \neq F(x_{adv}),~and~S_{Sim}(x,x_{adv}) \geq \epsilon
\end{align}
where $S_{Sim}(.)$ is a semantic and syntactic similarity function and $\epsilon$ is the minimum similarity between the input and adversarial samples. Here, $x_{adv} = x + \Delta x$ where $\Delta x$ is an imperceptible perturbation added into input sentence, $x$. 
The main challenge has been to come up with novel perturbation techniques for code-mixed data. Previous frameworks for monolingual data, language model can not be used to perform perturbations as such a model is currently not available for code-mixed data. In code-mixed data domain, synonym-based replacement strategy will not work as it is very difficult to derive synonyms from bilingual tokens and also their synonyms may have the undesired effect of changing the contextual meaning. Also, simply perturbing characters in a code-mixed token renders it meaningless in most of the cases, as the meaning as well as the semantic structure of a word might be disturbed. Hence, we had to take into account phonetic similarity in order to make token perturbations. We also had to identify language tags of each token correctly in order to perturb them into its complementary language.  
\section{Methodology}
In our proposed framework, we have developed a mechanism to attack text classification models under a black-box setting.
The goal of our framework is to identify the $k$ most important tokens $w_{1}$, $w_{2}$, ..., $w_{k}$ and apply a set of perturbation techniques, P = \{$p_{1}$, $p_{2}$, ..., $p_{m}$\} on them iteratively until we get $x_{adv}$ for the corresponding input sentence if attack is successful. More details of these different modules are given in following sub-sections. 

\subsection{Token Importance Calculation}
We consider a sample sentence of \textit{n} tokens, $X_i$ = \{$w_1$, $w_2$,..., $w_n$\}. In order to calculate the importance of a particular token in a sentence, we replace that particular token by an UNKNOWN token, and obtain the prediction vector for the modified sentence. 
Each token is assigned a score $S_{w_i}$ based on its impact on the sentence using a token importance calculation algorithm, and we select the top \textit{k} tokens as the set of \textit{most-important} tokens~($C_{CANDIDATES}$) that can be attacked. The scoring approach is undertaken so that the number of perturbations to the original sentence can be minimized. 
We calculate the token importance, $S_{w_i}$, using the equation below:
\begin{equation*}
\small
\label{eq:words_importance}
    S_{w_i}=
    \begin{cases}
      V_x(Y^{label})- V_{x-\{w_i\}}(Y^{pred})\,,\enspace \text{if}\ Y^{label}=Y^{pred} \\
      (V_x(Y^{label})-V_{x-\{w_i\}}(Y^{label}))\\ +\:(V_{x-\{w_i\}}(Y^{pred})-V_x(Y^{pred}))\,,\enspace\text{if}\ Y^{label}!=Y^{pred}
      
    \end{cases}
\end{equation*}
where, $V_x$ denotes the model prediction considering all the words in the original sentence, and, $V_{x-\{w_i\}}$ denotes the model prediction with token $w_i$ removed. $Y^{label}$ is the original label class, $Y^{pred}$ is the predicted class, and $V_x(Y^{label})$ denotes the probability value of the label class index in the model prediction.

\subsection{Perturbation Techniques}
Once we have our $C_{CANDIDATES}$, from $X_i$, we select tokens from this list in descending order of their importance in an iterative manner. Thereafter, we use a variety of perturbation techniques to alter this word in such a manner that it can adhere to the surrounding context and there is no significant change in the semantic structure of the sentence when this word is replaced. At the same time, we need to ensure that the perturbed word looks very much similar to a human-error and has a strong potential to force the target model to make a wrong prediction. 
We have used 3 perturbation techniques in our algorithm, in the order: (a) sub-word perturbation, (b) character-repetition, and (c) switch-word language. 
In order to identify a sub-word or character that can be perturbed within a token, we need to identify the language id of the token. We have different dictionaries defined for different languages, hence, language identification using the language id is really important so that we can load up the corresponding dictionary and replace the sub-word or character. We have used a character and phonetic based LSTM model~\cite{mandal2018language} to obtain language ids for various tokens present in a sentence.

\textbf{Sub-Word Perturbation :}
We have used a pre-existing dictionary of character groups that can be replaced by phonetically similar characters \cite{mandal2018language}. Essentially, these groups consists of character uni, bi and trigrams which are phonetically similar and are inter-changeably used in social media based on user backgrounds (e.g. \textit{pha} and \textit{f}, \textit{au} and \textit{ow}). Whenever such a character-group is present in any particular word in the given sentence, we replace it with its corresponding value(s) from the dictionary. For example, in Bengali, word 'bhalo'\textrightarrow'valo' (meaning good) and in Hindi, word 'gajab'\textrightarrow'gazb' (meaning surprising). 
The sub-word perturbation technique ensures that both the meaning as well as the semantic structure is preserved.

\textbf{Character-Repetition Perturbation :}
We also observed that character repetition was popular on social media, often to emphasize on something or for humour. Thus, we exploited this property and created a dictionary of top characters which are frequently repeated.
We select a character from the target word and repeat it once based on its value in the dictionary. Repeating certain characters do not alter the meaning of a word and also preserves its phonetic similarity, however, it might force a model to make a false prediction.
For example, in Hindi, word 'mafi'\textrightarrow'mafii' (meaning pardon) and in Bengali, word 'paoa'\textrightarrow'paooa' (meaning getting). 

\textbf{Switch-Word Language Perturbation :}
Given an input sentence, we have used a character and phonetic based LSTM model~\cite{mandal2018language} to obtain language ids for various tokens present in it. Once we have the language id of a token, we back-transliterate and translate the word to its complementary language using contextual information and an LSTM-based seq2seq model; for example, 'bacha'\textrightarrow'baby' and in Hindi, 'byaah'\textrightarrow'wedding'. This perturbation technique does not change the meaning of the word and also preserves contextual similarity to a great extent, however, it can force the model to make a false prediction.

\subsection{Iterative Inference}
In this step, we iteratively choose the next most-important word obtained using the token importance calculation algorithm, perturb it using the attack strategies one by one with subword-perturbation being used in the first trial, followed by character-repetition and switch-word language perturbations. The intuition behind the order of application of the above techniques is that the sub-word and letter-repetition perturbation techniques make changes only to a part of a word which tends to preserve its semantic structure the most. 
We obtain the model prediction vector, $F(X^{temp}_{adv})$ and the predicted class, $Y^{pred}_i$, by replacing each of these perturbed words in the original sentence. As soon as our prediction label becomes different from the original prediction of the unperturbed sentence, we claim that the attack is successful and terminate the process. If the system is unable to induce an attack even after trying out all possible perturbation techniques, we declare that an attack for the given sentence is unsuccessful. We also calculate the value of the maximum probability drop, $P_{drop}$, that has been induced in the label class.
We compute the label class probability value from the prediction vector of the adversarial sentence, $P^{temp}_{Y_i}$ that has produced the maximum drop among all perturbation techniques, for the current token, $w_i$, and the label class probability value, $P^{pred}_{Y_i}$ of the prediction vector obtained using the current perturbation function, and compute their maximum difference, $P^{ max}_{prob\_drop}$. $T_{max\_drop}$ is the perturbed token that produces the maximum probability drop in the label class of the prediction vector from ${X^{temp}_{adv}}$, obtained using $w_i$. The details of our attack framework is explained in Algorithm \ref{attack-algo}. 

\begin{algorithm}[]
\small
	\caption{CodeMixed Adversarial Attack}
	\label{h_algo}
	{\textbf{Input: }{Sentence $X_i=\{ w_1,w_2,...,w_n\}$, Set of token perturbation functions $f_{p_w}=\{f_{p_1},f_{p_2},f_{p_3}\}$, the corresponding ground truth label $Y^{label}_i$, target model $F$}, maximum number of words to perturb $k$.}\newline
    {\textbf{Output: }{Attack Success Flag, Predicted Label after attack, Probability Drop} }
\begin{algorithmic}[1]
  \State {Obtain model prediction vector, $F(X_i)$, with $X_i$}
  \State {Calculate  $P^{temp}_{Y_i}= F(X_i)[Y^{label}_i]$}
  \State {\textbf{Initialization}: $X_{\mathrm{adv}}\leftarrow X_i$}
  \For{each word $w_i$ in $X_i$}
  \State{Calculate the importance score $S_{w_i}$ of $w_i$ using Eq.~\eqref{eq:words_importance} }
  \EndFor
  \State {Obtain a final set of words T sorted by word importance scores $S{w_i}$.}
  \State {Select the top $k$ most important words using $S{w_i}$ from $T$ as $C_{CANDIDATES}$}
  \For{each word $w_{k}$ in $C_{CANDIDATES}$}
    \State {\textbf{Initialization}: $P^{max}_{prob\_drop}\leftarrow -1$}
    \State {\textbf{Initialization}: $T_{max\_drop}\leftarrow \texttt{None}$}
    \For{each perturbation technique $f_{p_i}$ from $f_{p_w}$}
      \State {Generate perturbed word $w^{p_i}_k$ of $w_k$, using $f_{p_i}$}
      \State{${X^{temp}_{adv}} = X_{adv}$ with $w_k$ replaced by $w^p_k$}
      \State{Obtain current model prediction with ${X^{temp}_{adv}}$, $F({X^{temp}_{adv}})$}
      \State{Calculate $Y^{pred}_i = argmax(F({X^{temp}_{adv}}))$}
      \State{$P^{pred}_{Y_i} = F({X^{temp}_{adv}})[Y^{label}_i]$}
      \If{$Y^{pred}_i != Y^{label}_i$}
        \Return {\texttt{True}, $Y^{pred}_i$}, $P^{ max}_{prob\_drop}$
      \Else
        \State{$P_{drop} = P^{temp}_{Y_i} - P^{pred}_{Y_i}$} 
        \If{$P_{drop} > P^{ max}_{prob\_drop}$}
          \State{$P^{ max}_{prob\_drop} = P_{drop}$}
          \State{$T_{max\_drop} = w^p_k$}
        \EndIf
      \EndIf
    \EndFor
    \State {$X_{adv} = X_{adv}$ with $w_k$ replaced by $T_{max\_drop}$}
    \State{$P^{temp}_{Y_i} = F(X_{adv})[Y_i]$} 
  \EndFor
  \Return{\texttt{False}, $Y^{label}_i$, $P^{ max}_{prob\_drop}$}

\end{algorithmic}
\label{attack-algo}
\end{algorithm}
\setlength{\textfloatsep}{0.1cm}

\begin{table*}[]
\small
\centering
\caption{Adversarial Attack Results On Different Models - Hindi-English Code-Mixed Data }
\label{hin-en_results}
\resizebox{\textwidth}{!}{%
\begin{tabular}{|c|c|c|c|c|c|c|c|c|c|c|c|c|c|} 
\hline
\multirow{3}{*}{\textbf{Model}} & \multirow{2}{*}{\begin{tabular}[c]{@{}c@{}}\textbf{Before }\\\textbf{Attack}\end{tabular}} & \multicolumn{12}{c|}{\textbf{After Attack}}                                                                                                                                                                                                                                                                                                                                                                         \\ 
\cline{3-14}
                                &                                                                                            & \multicolumn{4}{c|}{\textbf{Top 2 Words}}                                                                                & \multicolumn{4}{c|}{\textbf{Top 4 Words}}                                                                                                  & \multicolumn{4}{c|}{\textbf{Top 8 Words}}                                                                                                   \\ 
\cline{2-14}
                                & \textbf{F1}                                                                                & \textbf{F1} & \textbf{Time(s)} & \textbf{MOS} & $\mathbf{SR_{Adv}}$ & \textbf{F1} & \textbf{Time(s)} & \textbf{MOS} & $\mathbf{SR_{Adv}}$ & \textbf{F1} & \textbf{Time(s)} & \textbf{MOS} & $\mathbf{SR_{Adv}}$  \\ 
\hline
\textbf{Bi-LSTM-CNN}            & 0.8800                                                                                     & 0.5141      & 0.6504           & 0.1250       & 0.4678                                                                       & 0.3100      & 0.6641           & 0.1250       & 0.63                                                                                         & 0.2851      & 0.8267           & 0.4286       & 0.6901                                                                                          \\ 
\hline
\textbf{Bi-GRU-CNN}             & 0.9046                                                                                     & 0.5416      & 0.5922           & 0.2083       & 0.4436                                                                       & 0.3722      & 0.7015           & 0.2500       & 0.6129                                                                                         & 0.3024      & 0.7835           & 0.3496       & 0.687                                                                                          \\ 
\hline
\textbf{Transformer}            & 0.8736                                                                                     & 0.5579      & 0.3168           & 0.1250       & 0.4997                                                                       & 0.4500      & 0.3605           & 0.2917       & 0.5335                                                                                         & 0.3811      & 0.4096           & 0.8750       & 0.601                                                                                          \\ 
\hline
\textbf{char-CNN}               & 0.8708                                                                                     & 0.5441      & 0.5486           & 0.3750       & 0.4414                                                                       & 0.3948      & 0.6072           & 0.4150       & 0.5836                                                                                         & 0.3338      & 0.7144           & 0.5000       & 0.6543                                                                                          \\ 
\hline
\textbf{mBERT}                  & 0.8921                                                                                     & 0.7197      & 0.7984           & 0.2083       & 0.2673                                                                       & 0.5809      & 0.9766           & 0.9167       & 0.3974                                                                                         & 0.4820      & 1.9569           & 1.0417       & 0.4954                                                                                          \\
\hline
\end{tabular}
}
\end{table*}


\begin{table*}[]
\small
\centering
\caption{Adversarial Attack Results On Different Models - Bengali-English Code-Mixed Data}
\resizebox{\textwidth}{!}{%
\begin{tabular}{|c|c|c|c|c|c|c|c|c|c|c|c|c|c|} 
\hline
\multirow{3}{*}{ \textbf{Model} } & \multirow{2}{*}{\begin{tabular}[c]{@{}c@{}}\textbf{Before }\\\textbf{Attack} \end{tabular}} & \multicolumn{12}{c|}{\textbf{After Attack} } \\ 
\cline{3-14}
 &  & \multicolumn{4}{c|}{\textbf{Top 2 Words} } & \multicolumn{4}{c|}{\textbf{Top 4 Words} } & \multicolumn{4}{c|}{\textbf{Top 8 Words} } \\ 
\cline{2-14}
 & \textbf{F1}  & \textbf{F1} & \textbf{Time(s)}  & \textbf{MOS} & $\mathbf{SR_{Adv}}$ & \textbf{F1} & \textbf{Time(s)}  & \textbf{MOS} & $\mathbf{SR_{Adv}}$ & \textbf{F1} & \textbf{Time(s)}  & \textbf{MOS} & $\mathbf{SR_{Adv}}$ \\ 
\hline
\textbf{Bi-LSTM-CNN}  & 0.8966 & 0.7147 & 0.5052 & 0.2083 & 0.2778 & 0.4828 & 0.5938 & 0.2917 & 0.5122 & 0.3296 & 0.6967 & 0.4096 & 0.667 \\ 
\hline
\textbf{Bi-GRU-CNN}  & 0.8927 & 0.7255 & 0.5374 & 0.0833 & 0.263 & 0.5078 & 0.6410 & 0.4150 & 0.4836 & 0.3496 & 0.7628 & 0.7067 & 0.6819 \\ 
\hline
\textbf{Transformer}  & 0.8984 & 0.6852 & 0.3643 & 0.3750 & 0.3001 & 0.4726 & 0.4473 & 0.7083 & 0.5037 & 0.3149 & 0.5240 & 0.8333 & 0.6532 \\ 
\hline
\textbf{char-CNN}  & 0.8600 & 0.6185 & 0.4217 & 0.2917 & 0.4401 & 0.4957 & 0.4664 & 0.3750 & 0.4889 & 0.4191 & 0.5335 & 0.5000 & 0.562 \\ 
\hline
\textbf{mBERT}  & 0.9132 & 0.8365 & 0.6843 & 0.1428 & 0.158 & 0.7008 & 0.8821 & 0.4286 & 0.2768 & 0.5155 & 1.1967 & 0.5714 & 0.4369 \\
\hline
\end{tabular}
}
\label{bn-en_results}
\vspace{-4pt}
\end{table*}

\section{Experiments and Results}
We have used a sentiment classification task to demonstrate the capability of our framework. Initially, we had trained deep learning models on the given code-mixed sentiment classification datasets and evaluated their performance on the validation and test sets. We have used the same models to perform inference on the adversarial samples.
For models, we have taken into account several of the state-of-the-art models that have been used for sentiment classification over the years and have decided to finalize Bi-LSTM-CNN~\cite{jamatia2020deep}, Bi-GRU-CNN~\cite{jamatia2020deep}, Transformer~\cite{palomino2020palominoochoa}, char-CNN~\cite{zhang2015character} and mBERT~\cite{pires2019multilingual} based architectures for demonstration of the model agnostic nature of our adversarial attack technique.
The maximum input sequence length, vocabulary size, learning rate for these experiments were set at 25, 17k, and 0.001 respectively. We have summarized our results using F1-score, mean attack time per sentence, Mean Opinion Score (MOS)~\cite{streijl2016mean}, and Adversarial Attack Success Rate ($SR_{Adv}$) which is, the proportion of test data points on which adversarial attack has been successful, in Tables \ref{hin-en_results}, \ref{bn-en_results}. 

\textbf{Evaluation Metric : }
Here, we used Mean Opinion Score (MOS)~\cite{streijl2016mean} to evaluate our system. Cosine similarity is not applicable in this case as there is no pre-existing model to obtain the embeddings of code-mixed sentences. We were supported by a group of volunteers who had been provided a list of 100 questions each, for each one of the models. 
The MOS for each model is calculated by taking the average MOS given by different human participants involved in the study. Each participant was asked to enter a score in the range $0-4$ for each perturbed sentence among a list of perturbed sentences produced by that particular model, for a different configuration of the number of perturbed words; a $0$ indicates maximum similarity to the original sentence, while a $4$ indicates the least.

\textbf{Data Sets Details : }
We have used two code-mixed sentiment classification datasets on two different language pairs, ``\textit{Bn-En}" \cite{mandal2018preparing} and ``\textit{Hin-En}" \cite{patra2018sentiment} for our experiments to evaluate the effectiveness of our method. The ``\textit{Bn-En}" dataset contains 3206 and 943 samples in the training and test sets, while the ``\textit{Hin-En}" dataset contains 13845 and 1846 samples in training and test sets, respectively. The average length of sentences in ``\textit{Bn-En}", ``\textit{Hin-En}" are 15, 12, respectively, and the mean
Code-Mixing Index (CMI)~\cite{das-gamback-2014-identifying} for ``\textit{Bn-En}" and ``\textit{Hin-En}" datasets are  22.1, 18.57 respectively.

\textbf{Attacking Results and Ablation Study : }
From the results shown in Table~\ref{hin-en_results}, we can observe that the attack framework has been successful in forcing the models to make wrong predictions. Our approach is faster than other conventional approaches as those frameworks use another deep learning model or token similarity algorithm to bring about a successful perturbation. However, we use a hashing-based approach to partially or completely perturb tokens, which speeds up the entire operation to a huge extent.
mBERT and char-CNN have proven to be slightly more robust to adversarial attack than rest of the other models. In case of char-CNN, it can be attributed to the fact that there is no issue with characters being out-of-vocabulary; also, the words are represented by capturing information at the character level and the perturbation techniques affect only a certain fraction of characters of a word. Thus, there is a lesser scope of information loss in the token embedding vectors. The resistance of the mBERT model to adversarial attack can be justified on the grounds that it is pre-trained on a huge corpus of multilingual data, and the tokenization is done at the sub-word level, hence, perturbations that do not significantly alter the semantic structure and meaning of a sentence are robust to an attack. 
Table~\ref{bn-en_results} shows that the attack framework was successful in adversarial attacks on the given models, with mBERT and char-CNN remaining significantly more resilient to adversarial attacks than the other models.
The mBERT model remains the most robust model in this case as well.\\
We have also performed an experiment as ablation study to estimate the effectiveness of each of the 3 perturbation techniques by enumerating the corresponding perturbation success rate ($SR_{Perturb}$), which is the percentage of the vocabulary words that could be successfully perturbed using the given technique.
From Table \ref{ablation_study}, we infer that the letter-repetition technique is the most successful in Bengali-English data due to a possibility of fewer changes in the word semantic structure, and sub-word perturbation turns out to be the most effective in the Hindi-English data, which can be attributed to a greater sense of the semantic structure of the code-mixed tokens to changes at the sub-word level.
\begin{table}[]
\small
\centering
\caption{Code-mixed Perturbation Performance Study}
\begin{tabular}{|c|c|c|} 
\hline
\textbf{Dataset} & \textbf{Perturbation Methods} & $\mathbf{SR_{Perturb}}$ \\
\hline
\multirow{3}{*}{Bn-En} & Sub-Word & 85.9 \\ 
\cline{2-3}
 & Character-Repetition & 93.08 \\ 
\cline{2-3}
 & Switch-Word Language & 79.94 \\ 
\hline
\multirow{3}{*}{Hin-En} & Sub-Word & 88.5 \\ 
\cline{2-3}
 & Character-Repetition & 90.52 \\ 
\cline{2-3}
 & Switch-Word Language & 87 \\
\hline
\end{tabular}
\label{ablation_study}
\end{table}
\vspace{-6pt}
\section{Conclusion}
In this paper, we have presented a generic framework that can attack code-mixed classification models by identifying and perturbing important tokens. 
Our word-importance calculation algorithm ensures that an attack is successful with a very low percentage of word perturbations in the original sentence, and the entire process is completed within a very short duration. Also, the low values of MOS indicate that the perturbed sentences are very similar to the original ones. We have been able to reduce the F1-scores of both Bengali-English and Hindi-English code-mixed datasets, which shows that our attack framework can be successful with a variety of language pair. 

\bibliographystyle{ACM-Reference-Format}
\bibliography{sample-base}


\begin{thebibliography}{21}


\ifx \showCODEN    \undefined \def \showCODEN     #1{\unskip}     \fi
\ifx \showDOI      \undefined \def \showDOI       #1{#1}\fi
\ifx \showISBNx    \undefined \def \showISBNx     #1{\unskip}     \fi
\ifx \showISBNxiii \undefined \def \showISBNxiii  #1{\unskip}     \fi
\ifx \showISSN     \undefined \def \showISSN      #1{\unskip}     \fi
\ifx \showLCCN     \undefined \def \showLCCN      #1{\unskip}     \fi
\ifx \shownote     \undefined \def \shownote      #1{#1}          \fi
\ifx \showarticletitle \undefined \def \showarticletitle #1{#1}   \fi
\ifx \showURL      \undefined \def \showURL       {\relax}        \fi
\providecommand\bibfield[2]{#2}
\providecommand\bibinfo[2]{#2}
\providecommand\natexlab[1]{#1}
\providecommand\showeprint[2][]{arXiv:#2}

\bibitem[\protect\citeauthoryear{Das and Gamb{\"a}ck}{Das and
  Gamb{\"a}ck}{2014}]%
        {das-gamback-2014-identifying}
\bibfield{author}{\bibinfo{person}{Amitava Das} {and}
  \bibinfo{person}{Bj{\"o}rn Gamb{\"a}ck}.} \bibinfo{year}{2014}\natexlab{}.
\newblock \showarticletitle{Identifying Languages at the Word Level in
  Code-Mixed {I}ndian Social Media Text}. In
  \bibinfo{booktitle}{\emph{Proceedings of the 11th International Conference on
  Natural Language Processing}}. \bibinfo{publisher}{NLP Association of India},
  \bibinfo{address}{Goa, India}, \bibinfo{pages}{378--387}.
\newblock
\urldef\tempurl%
\url{https://www.aclweb.org/anthology/W14-5152}
\showURL{%
\tempurl}


\bibitem[\protect\citeauthoryear{Gao, Lanchantin, Soffa, and Qi}{Gao
  et~al\mbox{.}}{2018}]%
        {gao2018black}
\bibfield{author}{\bibinfo{person}{Ji Gao}, \bibinfo{person}{Jack Lanchantin},
  \bibinfo{person}{Mary~Lou Soffa}, {and} \bibinfo{person}{Yanjun Qi}.}
  \bibinfo{year}{2018}\natexlab{}.
\newblock \showarticletitle{Black-box generation of adversarial text sequences
  to evade deep learning classifiers}. In \bibinfo{booktitle}{\emph{2018 IEEE
  Security and Privacy Workshops (SPW)}}. IEEE, \bibinfo{pages}{50--56}.
\newblock


\bibitem[\protect\citeauthoryear{Jamatia, Swamy, Gamb{\"a}ck, Das, and
  Debbarma}{Jamatia et~al\mbox{.}}{2020}]%
        {jamatia2020deep}
\bibfield{author}{\bibinfo{person}{Anupam Jamatia}, \bibinfo{person}{Steve
  Swamy}, \bibinfo{person}{Bj{\"o}rn Gamb{\"a}ck}, \bibinfo{person}{Amitava
  Das}, {and} \bibinfo{person}{Swapam Debbarma}.}
  \bibinfo{year}{2020}\natexlab{}.
\newblock \showarticletitle{Deep Learning Based Sentiment Analysis in a
  Code-Mixed English-Hindi and English-Bengali Social Media Corpus}.
\newblock \bibinfo{journal}{\emph{International journal on artificial
  intelligence tools}} \bibinfo{volume}{29}, \bibinfo{number}{5}
  (\bibinfo{year}{2020}).
\newblock


\bibitem[\protect\citeauthoryear{Jin, Jin, Zhou, and Szolovits}{Jin
  et~al\mbox{.}}{2020}]%
        {jin2020bert}
\bibfield{author}{\bibinfo{person}{Di Jin}, \bibinfo{person}{Zhijing Jin},
  \bibinfo{person}{Joey~Tianyi Zhou}, {and} \bibinfo{person}{Peter Szolovits}.}
  \bibinfo{year}{2020}\natexlab{}.
\newblock \showarticletitle{Is bert really robust? a strong baseline for
  natural language attack on text classification and entailment}. In
  \bibinfo{booktitle}{\emph{Proceedings of the AAAI conference on artificial
  intelligence}}, Vol.~\bibinfo{volume}{34}. \bibinfo{pages}{8018--8025}.
\newblock


\bibitem[\protect\citeauthoryear{Li, Ji, Du, Li, and Wang}{Li
  et~al\mbox{.}}{2018}]%
        {li2018textbugger}
\bibfield{author}{\bibinfo{person}{Jinfeng Li}, \bibinfo{person}{Shouling Ji},
  \bibinfo{person}{Tianyu Du}, \bibinfo{person}{Bo Li}, {and}
  \bibinfo{person}{Ting Wang}.} \bibinfo{year}{2018}\natexlab{}.
\newblock \showarticletitle{Textbugger: Generating adversarial text against
  real-world applications}.
\newblock \bibinfo{journal}{\emph{arXiv preprint arXiv:1812.05271}}
  (\bibinfo{year}{2018}).
\newblock


\bibitem[\protect\citeauthoryear{Li, Ma, Guo, Xue, and Qiu}{Li
  et~al\mbox{.}}{2020}]%
        {li2020bert}
\bibfield{author}{\bibinfo{person}{Linyang Li}, \bibinfo{person}{Ruotian Ma},
  \bibinfo{person}{Qipeng Guo}, \bibinfo{person}{Xiangyang Xue}, {and}
  \bibinfo{person}{Xipeng Qiu}.} \bibinfo{year}{2020}\natexlab{}.
\newblock \showarticletitle{Bert-attack: Adversarial attack against bert using
  bert}.
\newblock \bibinfo{journal}{\emph{arXiv preprint arXiv:2004.09984}}
  (\bibinfo{year}{2020}).
\newblock


\bibitem[\protect\citeauthoryear{Liu, Chen, Feng, Wang, Ouyang, Sun, Huang, and
  Su}{Liu et~al\mbox{.}}{2020}]%
        {liu2020kk2018}
\bibfield{author}{\bibinfo{person}{Jiaxiang Liu}, \bibinfo{person}{Xuyi Chen},
  \bibinfo{person}{Shikun Feng}, \bibinfo{person}{Shuohuan Wang},
  \bibinfo{person}{Xuan Ouyang}, \bibinfo{person}{Yu Sun},
  \bibinfo{person}{Zhengjie Huang}, {and} \bibinfo{person}{Weiyue Su}.}
  \bibinfo{year}{2020}\natexlab{}.
\newblock \showarticletitle{Kk2018 at SemEval-2020 task 9: Adversarial training
  for code-mixing sentiment classification}.
\newblock \bibinfo{journal}{\emph{arXiv preprint arXiv:2009.03673}}
  (\bibinfo{year}{2020}).
\newblock


\bibitem[\protect\citeauthoryear{Madry, Makelov, Schmidt, Tsipras, and
  Vladu}{Madry et~al\mbox{.}}{2017}]%
        {madry2017towards}
\bibfield{author}{\bibinfo{person}{Aleksander Madry},
  \bibinfo{person}{Aleksandar Makelov}, \bibinfo{person}{Ludwig Schmidt},
  \bibinfo{person}{Dimitris Tsipras}, {and} \bibinfo{person}{Adrian Vladu}.}
  \bibinfo{year}{2017}\natexlab{}.
\newblock \showarticletitle{Towards deep learning models resistant to
  adversarial attacks}.
\newblock \bibinfo{journal}{\emph{arXiv preprint arXiv:1706.06083}}
  (\bibinfo{year}{2017}).
\newblock


\bibitem[\protect\citeauthoryear{Mandal, Das, and Das}{Mandal
  et~al\mbox{.}}{2018a}]%
        {mandal2018language}
\bibfield{author}{\bibinfo{person}{Soumil Mandal},
  \bibinfo{person}{Sourya~Dipta Das}, {and} \bibinfo{person}{Dipankar Das}.}
  \bibinfo{year}{2018}\natexlab{a}.
\newblock \showarticletitle{Language identification of bengali-english
  code-mixed data using character \& phonetic based lstm models}.
\newblock \bibinfo{journal}{\emph{arXiv preprint arXiv:1803.03859}}
  (\bibinfo{year}{2018}).
\newblock


\bibitem[\protect\citeauthoryear{Mandal, Mahata, and Das}{Mandal
  et~al\mbox{.}}{2018b}]%
        {mandal2018preparing}
\bibfield{author}{\bibinfo{person}{Soumil Mandal},
  \bibinfo{person}{Sainik~Kumar Mahata}, {and} \bibinfo{person}{Dipankar Das}.}
  \bibinfo{year}{2018}\natexlab{b}.
\newblock \showarticletitle{Preparing bengali-english code-mixed corpus for
  sentiment analysis of indian languages}.
\newblock \bibinfo{journal}{\emph{arXiv preprint arXiv:1803.04000}}
  (\bibinfo{year}{2018}).
\newblock


\bibitem[\protect\citeauthoryear{Meng and Chen}{Meng and Chen}{2017}]%
        {meng2017magnet}
\bibfield{author}{\bibinfo{person}{Dongyu Meng} {and} \bibinfo{person}{Hao
  Chen}.} \bibinfo{year}{2017}\natexlab{}.
\newblock \showarticletitle{Magnet: a two-pronged defense against adversarial
  examples}. In \bibinfo{booktitle}{\emph{Proceedings of the 2017 ACM SIGSAC
  conference on computer and communications security}}.
  \bibinfo{pages}{135--147}.
\newblock


\bibitem[\protect\citeauthoryear{Palomino and Ochoa-Luna}{Palomino and
  Ochoa-Luna}{2020}]%
        {palomino2020palominoochoa}
\bibfield{author}{\bibinfo{person}{Daniel Palomino} {and} \bibinfo{person}{Jose
  Ochoa-Luna}.} \bibinfo{year}{2020}\natexlab{}.
\newblock \bibinfo{title}{Palomino-Ochoa at SemEval-2020 Task 9: Robust System
  based on Transformer for Code-Mixed Sentiment Classification}.
\newblock
\newblock
\showeprint[arxiv]{2011.09448}~[cs.CL]


\bibitem[\protect\citeauthoryear{Patra, Das, and Das}{Patra
  et~al\mbox{.}}{2018}]%
        {patra2018sentiment}
\bibfield{author}{\bibinfo{person}{Braja~Gopal Patra},
  \bibinfo{person}{Dipankar Das}, {and} \bibinfo{person}{Amitava Das}.}
  \bibinfo{year}{2018}\natexlab{}.
\newblock \showarticletitle{Sentiment Analysis of Code-Mixed Indian Languages:
  An Overview of SAIL\_Code-Mixed Shared Task@ ICON-2017}.
\newblock \bibinfo{journal}{\emph{arXiv preprint arXiv:1803.06745}}
  (\bibinfo{year}{2018}).
\newblock


\bibitem[\protect\citeauthoryear{Patwa, Aguilar, Kar, Pandey, PYKL,
  Gamb{\"a}ck, Chakraborty, Solorio, and Das}{Patwa et~al\mbox{.}}{2020}]%
        {patwa2020semeval}
\bibfield{author}{\bibinfo{person}{Parth Patwa}, \bibinfo{person}{Gustavo
  Aguilar}, \bibinfo{person}{Sudipta Kar}, \bibinfo{person}{Suraj Pandey},
  \bibinfo{person}{Srinivas PYKL}, \bibinfo{person}{Bj{\"o}rn Gamb{\"a}ck},
  \bibinfo{person}{Tanmoy Chakraborty}, \bibinfo{person}{Thamar Solorio}, {and}
  \bibinfo{person}{Amitava Das}.} \bibinfo{year}{2020}\natexlab{}.
\newblock \showarticletitle{Semeval-2020 task 9: Overview of sentiment analysis
  of code-mixed tweets}.
\newblock \bibinfo{journal}{\emph{arXiv e-prints}} (\bibinfo{year}{2020}),
  \bibinfo{pages}{arXiv--2008}.
\newblock


\bibitem[\protect\citeauthoryear{Pires, Schlinger, and Garrette}{Pires
  et~al\mbox{.}}{2019}]%
        {pires2019multilingual}
\bibfield{author}{\bibinfo{person}{Telmo Pires}, \bibinfo{person}{Eva
  Schlinger}, {and} \bibinfo{person}{Dan Garrette}.}
  \bibinfo{year}{2019}\natexlab{}.
\newblock \showarticletitle{How multilingual is multilingual bert?}
\newblock \bibinfo{journal}{\emph{arXiv preprint arXiv:1906.01502}}
  (\bibinfo{year}{2019}).
\newblock


\bibitem[\protect\citeauthoryear{Ren, Deng, He, and Che}{Ren
  et~al\mbox{.}}{2019}]%
        {ren2019generating}
\bibfield{author}{\bibinfo{person}{Shuhuai Ren}, \bibinfo{person}{Yihe Deng},
  \bibinfo{person}{Kun He}, {and} \bibinfo{person}{Wanxiang Che}.}
  \bibinfo{year}{2019}\natexlab{}.
\newblock \showarticletitle{Generating natural language adversarial examples
  through probability weighted word saliency}. In
  \bibinfo{booktitle}{\emph{Proceedings of the 57th annual meeting of the
  association for computational linguistics}}. \bibinfo{pages}{1085--1097}.
\newblock


\bibitem[\protect\citeauthoryear{Sridhar and Sridhar}{Sridhar and
  Sridhar}{1980}]%
        {sridhar1980syntax}
\bibfield{author}{\bibinfo{person}{Shikaripur~N Sridhar} {and}
  \bibinfo{person}{Kamal~K Sridhar}.} \bibinfo{year}{1980}\natexlab{}.
\newblock \showarticletitle{The syntax and psycholinguistics of bilingual code
  mixing.}
\newblock \bibinfo{journal}{\emph{Canadian Journal of Psychology/Revue
  canadienne de psychologie}} \bibinfo{volume}{34}, \bibinfo{number}{4}
  (\bibinfo{year}{1980}), \bibinfo{pages}{407}.
\newblock


\bibitem[\protect\citeauthoryear{Streijl, Winkler, and Hands}{Streijl
  et~al\mbox{.}}{2016}]%
        {streijl2016mean}
\bibfield{author}{\bibinfo{person}{Robert~C Streijl}, \bibinfo{person}{Stefan
  Winkler}, {and} \bibinfo{person}{David~S Hands}.}
  \bibinfo{year}{2016}\natexlab{}.
\newblock \showarticletitle{Mean opinion score (MOS) revisited: methods and
  applications, limitations and alternatives}.
\newblock \bibinfo{journal}{\emph{Multimedia Systems}} \bibinfo{volume}{22},
  \bibinfo{number}{2} (\bibinfo{year}{2016}), \bibinfo{pages}{213--227}.
\newblock


\bibitem[\protect\citeauthoryear{Sun}{Sun}{2020}]%
        {sun2020natural}
\bibfield{author}{\bibinfo{person}{Lichao Sun}.}
  \bibinfo{year}{2020}\natexlab{}.
\newblock \showarticletitle{Natural backdoor attack on text data}.
\newblock \bibinfo{journal}{\emph{arXiv preprint arXiv:2006.16176}}
  (\bibinfo{year}{2020}).
\newblock


\bibitem[\protect\citeauthoryear{Tan and Joty}{Tan and Joty}{2021}]%
        {tan2021code}
\bibfield{author}{\bibinfo{person}{Samson Tan} {and} \bibinfo{person}{Shafiq
  Joty}.} \bibinfo{year}{2021}\natexlab{}.
\newblock \showarticletitle{Code-Mixing on Sesame Street: Dawn of the
  Adversarial Polyglots}.
\newblock \bibinfo{journal}{\emph{arXiv preprint arXiv:2103.09593}}
  (\bibinfo{year}{2021}).
\newblock


\bibitem[\protect\citeauthoryear{Zhang, Zhao, and LeCun}{Zhang
  et~al\mbox{.}}{2015}]%
        {zhang2015character}
\bibfield{author}{\bibinfo{person}{Xiang Zhang}, \bibinfo{person}{Junbo Zhao},
  {and} \bibinfo{person}{Yann LeCun}.} \bibinfo{year}{2015}\natexlab{}.
\newblock \showarticletitle{Character-level convolutional networks for text
  classification}.
\newblock \bibinfo{journal}{\emph{arXiv preprint arXiv:1509.01626}}
  (\bibinfo{year}{2015}).
\newblock


\end{thebibliography}

\pagebreak
\appendix

\section{Appendix}
\label{sec:appendix}

\section{Qualitative Results}
In Tables \ref{bengali_attack_samples} and \ref{hindi_attack_samples}, we show the qualitative results of the effect of our perturbation algorithm on both the Bengali-English and Hindi-English code-mixed datasets. We observe that the perturbations that have been introduced in the sentences are negligible when the number of perturbed tokens are lesser, which can be attributed to an unintentional human error or a typing error. However, as the number of tokens perturbed increases, the attack success rate increases while the quality of sentences degrade. The perturbations do not alter the overall meaning of a sentence, neither does it effect the semantic structure. Hence, our perturbation algorithm can attack deep learning models in a very subtle manner.

\section{Error Analysis}
We observe that it is relatively easy to produce an adversarial attack in the case of shorter sentences. It can be partially explained by the fact that in many cases, a particular word might have a much higher contribution to the overall context of a short sentence compared to a longer sentence. Hence, we are easily able to switch the predicted label for a shorter sentence after perturbing less than four tokens only, while we might need to perturb a larger number of tokens in the case of a longer sentence.
Also, some perturbations like the letter repetition perturbation might add a larger number of redundant characters to a word which might be quite conspicuous, however, they do not alter the meaning of the sentence and is unable to "fool" the model because of its phonetic similarity. We have provided some unsuccessful adversarial samples on mBERT model for both Bengali-English and Hindi-English datasets in Tables \ref{bengali_attack_error_samples} and \ref{hindi_attack_error_samples}.

\begin{table*}[]
\small
\centering
\caption{Hindi-English Adversarial Attack Samples}
\label{hindi_attack_samples}
\resizebox{\textwidth}{!}{%
\begin{tabular}{|c|c|c|c|c|c|c|} 
\hline
\multicolumn{7}{|c|}{\begin{tabular}[c]{@{}c@{}}\textbf{Input Text :} pick up da cal damn dumb gal he neds u dhaniya pyar krti h \\tu use bat kr use mar rha hoga wo andar se\end{tabular}} \\ 
\hhline{|=======|}
\textbf{Model} & \begin{tabular}[c]{@{}c@{}}\textbf{Perturbation}\\\textbf{Level}\\\textbf{(K)}\end{tabular} & \textbf{Examples} & \begin{tabular}[c]{@{}c@{}}\textbf{Ground }\\\textbf{Truth}\end{tabular} & \begin{tabular}[c]{@{}c@{}}\textbf{Predicted }\\\textbf{Label}\end{tabular} & \begin{tabular}[c]{@{}c@{}}\textbf{Perturbed }\\\textbf{Label}\end{tabular} & \textbf{MOS} \\ 
\hline
\multirow{3}{*}{\begin{tabular}[c]{@{}c@{}}\textbf{Bi-LSTM}\\\textbf{CNN}\end{tabular}} & 2 & \begin{tabular}[c]{@{}c@{}}piiik up da cal damn dumb gal he neds u dhaniya pyar krti h \\tu use bat kr use mar rha hoga wo andar se\end{tabular} & \multirow{3}{*}{neutral} & \multirow{3}{*}{neutral} & neutral & 0.125 \\ 
\cline{2-3}\cline{6-7}
 & 4 & \begin{tabular}[c]{@{}c@{}}pick up da call damn dumb gal he neds u dhaniya pyar krti h\\~tu use bat kr use mar rha hoga wo andar se\end{tabular} &  &  & neutral & 0 \\ 
\cline{2-3}\cline{6-7}
 & 8 & \begin{tabular}[c]{@{}c@{}}pick up da cal damn dumb gal he neds u dhaniya pyar krti h \\tu use bat kr use mar rha haaga wo within se\end{tabular} &  &  & neutral & 0.125 \\ 
\hline
\multirow{3}{*}{\begin{tabular}[c]{@{}c@{}}\textbf{Bi-GRU}\\\textbf{CNN}\end{tabular}} & 2 & \begin{tabular}[c]{@{}c@{}}pick up da cal damn dumb gaal he neds u dhaniya pyar krti h \\tu use talk kr use mar rha hoga wo andar se\end{tabular} & \multirow{3}{*}{neutral} & \multirow{3}{*}{neutral} & neutral & 0.125 \\ 
\cline{2-3}\cline{6-7}
 & 4 & \begin{tabular}[c]{@{}c@{}}pick up da cal damn dumb gaal he neds u dhniya pyar krti h \\tu use talk kr use mar rha hoga wo andar se\end{tabular} &  &  & negative & 0.125 \\ 
\cline{2-3}\cline{6-7}
 & 8 & \begin{tabular}[c]{@{}c@{}}pick up da cal damn dumb glll he neds u dhniya pyar krti h \\tu use talk kr use mar rha hoga wo andar se\end{tabular} &  &  & negative & 0.150 \\ 
\hline
\multirow{3}{*}{\textbf{charCNN}} & 2 & \begin{tabular}[c]{@{}c@{}}pick up da cal damn dumb gal he neees u dhaniya pyar krti h \\tu use bat kr use mar rha hoga wo andr se\end{tabular} & \multirow{3}{*}{neutral} & \multirow{3}{*}{neutral} & negative & 0.125 \\ 
\cline{2-3}\cline{6-7}
 & 4 & \begin{tabular}[c]{@{}c@{}}pick up da cal dmn dumb gal he neds u dhaniya pyar krti h \\tu use bat kr use mar rha hoga wo andr se\end{tabular} &  &  & neutral & 0.125 \\ 
\cline{2-3}\cline{6-7}
 & 8 & \begin{tabular}[c]{@{}c@{}}pick up daa clll dmn dumb gal he neds u dhaniya pyar krti h \\tu use bat kr use mar rha hooga wo andaarrrr se\end{tabular} &  &  & negative & 0.250 \\ 
\hline
\multirow{3}{*}{\textbf{Transformer}} & 2 & \begin{tabular}[c]{@{}c@{}}pick up da cal damn dumb ladki he neds pyar krti \\tu use bat kr use mar rha hoga wo andar se\end{tabular} & \multirow{3}{*}{neutral} & \multirow{3}{*}{neutral} & neutral & 0.125 \\ 
\cline{2-3}\cline{6-7}
 & 4 & \begin{tabular}[c]{@{}c@{}}pick up da cal damn duumb ladki he neds pyar krti \\tu use bat kr use mar rha hoga wo andar se\end{tabular} &  &  & positive & 0.125 \\ 
\cline{2-3}\cline{6-7}
 & 8 & \begin{tabular}[c]{@{}c@{}}pick up da cal damn goonga ladki he neds u dhaniya pyar krti h \\tu use bat kr use mar rha hoga wo andar se\end{tabular} &  &  & positive & 0.125 \\ 
\hline
\multirow{3}{*}{\textbf{mBERT}} & 2 & \begin{tabular}[c]{@{}c@{}}pick up da cal damn dumb gal hee neds dhaniya pyaarr krti \\tu use bat kr use mar rha hoga wo andar se\end{tabular} & \multirow{3}{*}{neutral} & \multirow{3}{*}{neutral} & neutral & 0.125 \\ 
\cline{2-3}\cline{6-7}
 & 4 & \begin{tabular}[c]{@{}c@{}}pick up da cal damn dumb gal hee neds dhaniya pyaarr krti \\tu use bat kr use mar rha hoga wo andar se\end{tabular} &  &  & neutral & 0.125 \\ 
\cline{2-3}\cline{6-7}
 & 8 & \begin{tabular}[c]{@{}c@{}}pick up da cal damn dumb gal Hee nedee dhaniya paaarr krtii \\tu useeee bat kr useeee mar rha hoga wo insiderr se\end{tabular} &  &  & negative & 0.375 \\
\hline
\end{tabular}
}
\end{table*}

\begin{table*}[]
\small
\centering
\caption{Bengali-English Adversarial Attack Samples}
\label{bengali_attack_samples}
\resizebox{\textwidth}{!}{%
\begin{tabular}{|c|c|c|c|c|c|c|} 
\hline
\multicolumn{7}{|c|}{\textbf{Input Text :}~ vai eita hobe na vai tmio jao vai plz vai vipode pore jabo} \\ 
\hhline{|=======|}
\textbf{Model} & \begin{tabular}[c]{@{}c@{}}\textbf{Perturbation }\\\textbf{Level}\\\textbf{(K)}\end{tabular} & \textbf{Examples} & \begin{tabular}[c]{@{}c@{}}\textbf{Ground }\\\textbf{Truth}\end{tabular} & \begin{tabular}[c]{@{}c@{}}\textbf{Predicted }\\\textbf{Label}\end{tabular} & \begin{tabular}[c]{@{}c@{}}\textbf{Perturbed }\\\textbf{Label}\end{tabular} & \textbf{MOS} \\ 
\hline
\multirow{3}{*}{\begin{tabular}[c]{@{}c@{}}\textbf{Bi-LSTM}\\\textbf{CNN}\end{tabular}} & 2 & vai eeita hobe na vai tmio jao vai plz vai vipoode pore jabo & \multirow{3}{*}{negative} & \multirow{3}{*}{negative} & neutral & 0.125 \\ 
\cline{2-3}\cline{6-7}
 & 4 & vai eeita hobe na vai tmio jao vai plz vai vipoode pore jabo &  &  & neutral & 0.125 \\ 
\cline{2-3}\cline{6-7}
 & 8 & vai eeeta hobe na vai tmio jao vai plz vai veepode pore jabo &  &  & neutral & 0.150 \\ 
\hline
\multirow{3}{*}{\begin{tabular}[c]{@{}c@{}}\textbf{Bi-GRU}\\\textbf{CNN}\end{tabular}} & 2 & vai eeeta hobe na vai youtoo jao vai plz vai vipode pore jabo & \multirow{3}{*}{negative} & \multirow{3}{*}{negative} & negative & 0.150 \\ 
\cline{2-3}\cline{6-7}
 & 4 & vai eeeta hobe na vai tmeeo jao vai plz vai veepode pore jabo &  &  & neutral & 0.150 \\ 
\cline{2-3}\cline{6-7}
 & 8 & vai eit hobe na vai tmeeo jao vai plz vai vipoode pore jabo &  &  & neutral & 0.150 \\ 
\hline
\multirow{3}{*}{\textbf{charCNN}} & 2 & voi eita hobe na voi tmio jao voi pllz voi vipode pore jbo & \multirow{3}{*}{negative} & \multirow{3}{*}{negative} & neutral & 0.125 \\ 
\cline{2-3}\cline{6-7}
 & 4 & voi eita hobe na voi tmio jao voi pllz voi vipode pore jbo &  &  & neutral & 0.125 \\ 
\cline{2-3}\cline{6-7}
 & 8 & vaee eita hobe na vaee tmio jao vaee plzz vaee vipode pore jaaboo &  &  & neutral & 0.125 \\ 
\hline
\multirow{3}{*}{\textbf{Transformer}} & 2 & vai eeeta hobe na vai tmio jao vai plz vai vipoode pore jabo & \multirow{3}{*}{negative} & \multirow{3}{*}{negative} & neutral & 0.125 \\ 
\cline{2-3}\cline{6-7}
 & 4 & vai eit hobe na vai tmio jao vai plz vai vipoode pore jabo &  &  & neutral & 0.125 \\ 
\cline{2-3}\cline{6-7}
 & 8 & vai eeita hobe na vai tmio jao vai plz vai vipoode pore jabo &  &  & neutral & 0.125 \\ 
\hline
\multirow{3}{*}{\textbf{mBERT}} & 2 & vai eeittt hobe na vai tmio jao vai plz vai vipoode pore jabo & \multirow{3}{*}{negative} & \multirow{3}{*}{negative} & negative & 0.150 \\ 
\cline{2-3}\cline{6-7}
 & 4 & vai eeeta hobe na vai tmio jao vai plz vai vipoode pore jabo &  &  & negative & 0.125 \\ 
\cline{2-3}\cline{6-7}
 & 8 & vai eit hobe n vai tmio jaaao vai pzzz vai vipodeeee p0re xbo &  &  & positive & 0.200 \\
\hline
\end{tabular}%
}
\end{table*}

\begin{table*}[]
\small
\centering
\caption{Bengali-English Adversarial Attack Error Samples}
\label{bengali_attack_error_samples}
\resizebox{\textwidth}{!}{%
\begin{tabular}{|c|c|c|c|c|c|} 
\hline
\multicolumn{6}{|c|}{\begin{tabular}[c]{@{}c@{}}\textbf{Input Text : }Denmark er movie Festen conventional film theke \\ekdom alada Eta bohu festival prize peyechilo\end{tabular}} \\ 
\hline
\begin{tabular}[c]{@{}c@{}}\textbf{Perturbation}\\\textbf{Level}\\\textbf{(k)}\end{tabular} & \textbf{Examples } & \textbf{Ground Truth } & \textbf{Predicted Label } & \textbf{Perturbed Label } & \textbf{MOS } \\ 
\hline
2 & \begin{tabular}[c]{@{}c@{}}Denmark err movie Festen conventional film theke \\ekdom alda Eta bohu festival prize peyechilo\end{tabular} & \multirow{3}{*}{negative} & \multirow{3}{*}{negative} & negative & 0.125 \\ 
\cline{1-2}\cline{5-6}
4 & \begin{tabular}[c]{@{}c@{}}Denmark err movie Festen conventional film theke \\ekdom alad Eta bohu festival prize peyechilo\end{tabular} &  &  & negative & 0.125 \\ 
\cline{1-2}\cline{5-6}
8 & \begin{tabular}[c]{@{}c@{}}Denmark errr muvie Festen conventttonal film theke \\ekdom alad Eaaa bohw festival prije piiyechiloo\end{tabular} &  &  & negative & 3.125 \\
\hline
\end{tabular}
}
\end{table*}

\begin{table*}[]
\small
\centering
\caption{Hindi-English Adversarial Attack Error Samples}
\label{hindi_attack_error_samples}
\resizebox{\textwidth}{!}{%
\begin{tabular}{|c|c|c|c|c|c|} 
\hline
\multicolumn{6}{|c|}{\textbf{Input Text : }Hlo salman sir mai aur mere mami papa apke bhut bde fan h} \\ 
\hline
\begin{tabular}[c]{@{}c@{}}\textbf{Perturbation}\\\textbf{~Level}\\\textbf{(k)}\end{tabular} & \textbf{Examples } & \textbf{Ground Truth } & \textbf{Predicted Label } & \textbf{Perturbed Label } & \textbf{MOS } \\ 
\hline
2 & Hlo salman sir mai aur mere mami ppa apke bhut bde fan h & neutral & neutral & neutral & 0.125 \\ 
\hline
4 & Hlo salman sir mai aur mere mami ppa apke bhut bigee fan h & neutral & neutral & neutral & 0.125 \\ 
\hline
8 & Hlo salman sir mai aur mere mami ppa apke bhut bigee fnnn h & neutral & neutral & neutral & 0.675 \\
\hline
\end{tabular}
}
\end{table*}

\end{document}